\documentclass[sigconf]{acmart}

\usepackage{CJKutf8}
\usepackage{booktabs}
\usepackage{graphicx}
\usepackage{array}
\usepackage{bm}
\usepackage{amsmath}
\renewcommand{\abstractname}{\MakeUppercase{Abstract}}
\renewcommand{\refname}{\MakeUppercase{References}}
\usepackage{titlesec}

\titleformat{\section}[block]{\normalfont\Large\bfseries}{\thesection}{1em}{\MakeUppercase}
\titleformat{\subsection}[block]{\normalfont\large\bfseries}{\thesubsection}{1em}{\MakeUppercase}

\usepackage{balance}
\usepackage{threeparttable}
\usepackage{multirow}
\usepackage{makecell}

\usepackage{latexsym}
\newcommand{\tabincell}[2]{\begin{tabular}{@{}#1@{}}#2\end{tabular}}
\usepackage[T1]{fontenc}
\usepackage[utf8]{inputenc}
\usepackage{float}

\usepackage{microtype}

\usepackage{xcolor}

\newcommand{\eat}[1]{{}}

\def\BibTeX{{\rm B\kern-.05em{\sc i\kern-.025em b}\kern-.08em
    T\kern-.1667em\lower.7ex\hbox{E}\kern-.125emX}}
\usepackage{cite}
\usepackage{amsmath,amssymb,amsfonts}
\usepackage{algorithmic}
\usepackage[ruled,linesnumbered,noline,noend]{algorithm2e}

\SetCommentSty{mycommfont}

\SetNlSty{}{}{}

\let\oldnl\nl% Store \nl in \oldnl
\newcommand\nonl{%
  \renewcommand{\nl}{\let\nl\oldnl}}% Remove line number for one line
  
\usepackage{hyperref}
\hypersetup{
    colorlinks=true,
    linkcolor=blue,}
\AtBeginDocument{%
  \providecommand\BibTeX{{%
    \normalfont B\kern-0.5em{\scshape i\kern-0.25em b}\kern-0.8em\TeX}}}

\copyrightyear{2023}
\acmYear{2023}
\setcopyright{acmlicensed}\acmConference[CIKM '23]{Proceedings of the 32nd ACM International Conference on Information and Knowledge Management}{October 21--25, 2023}{Birmingham, United Kingdom}
\acmBooktitle{Proceedings of the 32nd ACM International Conference on Information and Knowledge Management (CIKM '23), October 21--25, 2023, Birmingham, United Kingdom}
\acmPrice{15.00}
\acmDOI{10.1145/3583780.3614904}
\acmISBN{979-8-4007-0124-5/23/10}

\begin{document}

%\title{Attention-Adapter: A Parameter-Efficient Adapter Tuning Method for Pre-trained Language Models}%Attention-based
% \title{0.03\%! An Extreme Parameter-Efficient Adapter Tuning Method for Pre-trained Language Models}
\title{Hadamard Adapter: An Extreme Parameter-Efficient Adapter Tuning Method for Pre-trained Language Models}

\author{Yuyan Chen}
\authornote{Work done while this author was an intern at Microsoft Research.}
\email{chenyuyan21@m.fudan.edu.cn}
\orcid{0000-0002-4381-486X}
\affiliation{%
  \institution{Shanghai Key Laboratory of Data Science, School of Computer Science, Fudan University}
  \city{Shanghai}
  \country{China}
}

\author{Qiang Fu}
\authornote{The corresponding authors.}
\email{qifu@microsoft.com}
\orcid{0000-0002-5821-7267}
\affiliation{%
  \institution{Microsoft}
  \city{Beijing}
  \country{China}
}

\author{Ge Fan}
\email{ge.fan@outlook.com}
\orcid{0000-0001-5653-1626}
\affiliation{%
  \institution{Tencent}
  \city{Shenzhen}
  \country{China}
}

\author{Lun Du}
\email{ludu@microsoft.com}
\orcid{0000-0002-7625-0650}
\affiliation{%
  \institution{Microsoft}
  \city{Beijing}
  \country{China}
}

\author{Jian-Guang Lou}
\email{jlou@microsoft.com}
\orcid{0000-0001-8496-033X}
\affiliation{%
  \institution{Microsoft}
  \city{Beijing}
  \country{China}
}

\author{Shi Han}
\email{shihan@microsoft.com}
\orcid{0000-0002-0360-6089}
\affiliation{%
  \institution{Microsoft}
  \city{Beijing}
  \country{China}
}

\author{Dongmei Zhang}
\email{dongmeiz@microsoft.com}
\orcid{0000-0002-9230-2799}
\affiliation{%
  \institution{Microsoft}
  \city{Beijing}
  \country{China}
}

\author{Zhixu Li}
\authornotemark[2]
\email{zhixuli@fudan.edu.cn}
\orcid{0000-0003-2355-288X}
\affiliation{%
  \institution{Shanghai Key Laboratory of Data Science, School of Computer Science, Fudan University}
  \city{Shanghai}
  \country{China}
}

\author{Yanghua Xiao}
\authornotemark[2]
\email{shawyh@fudan.edu.cn}
\orcid{0000-0001-8403-9591}
\affiliation{%
  \institution{Shanghai Key Laboratory of Data Science, School of Computer Science, Fudan University, Fudan-Aishu Cognitive Intelligence Joint Research Center}
  \city{Shanghai}
  \country{China}
}

% \author{Anonymous et al.}

%\settopmatter{printacmref=false} % Removes citation information below abstract
% \renewcommand\footnotetextcopyrightpermission[1]{} % removes footnote with conference information in first column
%\renewcommand{\shortauthors}{Chen et al.}
\renewcommand{\shortauthors}{Yuyan Chen et al.}
% \pagestyle{plain} % removes running headers
\begin{abstract}
Recent years, Pre-trained Language models (PLMs) have swept into various fields of artificial intelligence and achieved great success.
However, most PLMs, such as T5 and GPT3, have a huge amount of parameters, fine-tuning them is often expensive and time consuming, and storing them takes up a lot of space.
Therefore, it is necessary to adopt a parameter-efficient approach to reduce parameters of PLMs in fine-tuning without compromising their performance in downstream tasks.
In this paper, we design a novel adapter which only acts on self-attention outputs in PLMs. This adapter adopts element-wise linear transformation using Hadamard product, hence named as Hadamard adapter, requires the fewest parameters compared to previous parameter-efficient adapters.
In addition, we also summarize some tuning patterns for Hadamard adapter shared by various downstream tasks, expecting to provide some guidance for further parameter reduction with shared adapters in future studies.
The experiments conducted on the widely-used GLUE benchmark with several SOTA PLMs prove that the Hadamard adapter achieves competitive performance with only 0.033\% parameters compared with full fine-tuning, and it has the fewest parameters compared with other adapters. Moreover, we further find that there is also some redundant layers in the Hadamard adapter which can be removed to achieve more parameter efficiency with only 0.022\% parameters.
\end{abstract}



%    <concept>
%        <concept_id>10010147.10010178.10010179</concept_id>
%        <concept_desc>Computing methodologies~Natural language processing</concept_desc>
%        <concept_significance>500</concept_significance>
%        </concept>
%  </ccs2012>

% \ccsdesc[500]{Computing methodologies~Natural language processing}

% \keywords{Humor evaluation framework, Humor understanding, Chinese humor dataset, Pre-trained language models}


\maketitle

% \vspace{-0.5em}
\section{Introduction}
% \vspace{-0.5em}
Recent years, Pre-trained Language models (PLMs) have swept into various fields of artificial intelligence and achieved great success~\citep{chen2024temporalmed,chen2023can,chen2023hallucination,chen2022grow,chen2024talk,chen2023xmqas,chen2023mapo}.
%
%Pre-trained Language models (PLMs) are widely used in Natural Language Processing (NLP), such as question-answering systems, searching engine, etc. 
%Most PLMs, such as T5, GPT3, have a large amount of parameters which makes them expensive and time-consuming to train from scratch and space-consuming to store.The mainstream paradigm for solving downstream tasks with PLMs is to make fine-tuning. Although fine-tuning is faster and need fewer computing resources compared with training from scratch, we still suppose that some parameters in the fine-tuning are redundant which can be removed or replaced with a simpler network to achieve parameter-efficiency.Therefore, it's important to adopt parameter-efficient methods to reduce parameters but maintain similar performance of PLMs on downstream tasks.
The mainstream paradigm for adapting PLMs to downstream tasks is fine-tuning. As most PLMs, such as T5~\citep{T5}, GPT3~\citep{GPT3}, have a large amount of parameters, fine-tuning them is often expensive and time consuming, and storing them takes up a lot of space.
%
%it is usually expensive and time-consuming to fine-tuning them and space-consuming to store them. %The overhead of storing models is severe especially when dealing with multiple downstream tasks. 
%
It has been revealed that there are a lot of redundant parameters in the process of fine-tuning~\citep{A_Survey_of,Large_Scale_Distributed}.
%In addition, previous research works~\citep{A_Survey_of,Large_Scale_Distributed} have shown that there is big redundancy among such great number of parameters during fine-tuning. 
%
Thus, it is necessary to greatly reduce the scale of parameters in fine-tuning without compromising PLMs' performance in downstream tasks.
%Therefore, it's important to adopt parameter-efficient methods to reduce parameters need to be learned/changed while maintaining similar performance of PLMs on downstream task, which can reduce both the training computational cost and the model storage cost. 

Previous parameter-efficient fine-tuning of PLMs mainly contains three categories of methods, i.e., adapter tuning, prefix tuning, and prompt tuning. Adapter tuning~\citep{Parameter-Efficient_Transfer_Learning} is to inject a small neural network module into each or some layers of the PLMs. During fine-tuning, only the parameters of this small module need to be learned. It has promising performance in NLP, which achieves comparable performance with fine-tuning while adding no more than 4\% task-specific parameters~\citep{Parameter-Efficient_Transfer_Learning,Exploring_versatile_generative}.
Prefix tuning~\citep{Prefix-Tuning} and prompt tuning~\citep{The_Power_of} preset additional adjustable prefix tokens in the input or hidden layer, and only these soft prompts are trained during the fine-tuning of downstream tasks. 
%
%To improve the efficiency of fine-tuning, the existing methods mainly contain model compression~\citep{A_Survey_of}, and and structure optimization~\citep{???}.
%
%Model compression includes knowledge distillation~\citep{Do_deep_nets,Distilling_the_Knowledge}, which trains small models with the knowledge learned by large models, so that small models have the generalization ability of large models; and quantization~\citep{Compressing_Deep_Convolutional,Quantized_Convolutional_Neural}, which reduces the accuracy of large models.
%
In addition to the above three parameter-efficient fine-tuning ways, the existing efforts also works on model compression~\citep{A_Survey_of}, including knowledge distillation~\citep{Do_deep_nets,Distilling_the_Knowledge}, which transfers the knowledge learned by large models to small models, such that small models can have the generalization ability of large models;
quantization~\citep{Compressing_Deep_Convolutional,Quantized_convolutional_neural}, which reduces the accuracy of large models within the acceptable range;
pruning~\citep{Data-free_parameter_pruning,Neural_and_Evolutionary}, which removes less useful connections in the model; and structure optimization, such as matrix decomposition~\citep{Convolutional_neural_networks}, parameter sharing~\citep{Soft_Weight-Sharing_for}, etc.
%}
%However, although the current endeavors achieve competitive performance of downstream tasks with much fewer parameters, we consider that parameter efficiency still has room to improve.
%
However, although the current endeavors achieve competitive performance in downstream tasks with much fewer parameters, we believe there is still room for improvement in parametric efficiency.

It's well-known that the attention mechanism, especially self-atten-tion, is one of the core modules that enable PLMs achieve superior performance in various downstream tasks~\citep{Attention_Is_All}.
%
%It directly establishes the relationship between any two words in a sentence which can model both short-distance and long-distance contextual dependency well.
%
%Considering the importance of the self attention, we assume the attention modules also play important roles in fine-tuning on the downstream tasks. %
%Therefore, we design a novel kernel adapter based on attention output to make parameter-efficient tuning for PLMs. Inspired by kernel function in Support Vector Machine (SVM), which is used to reduce computational complexity in high-dimensional space, our proposed kernel adapter is to inject a simpler feedforward neural network (FFN) on the output of self-attention to approximate fine-tuning performance. 
%
%According to our empirical analysis, we adopt linear fitting, instead of high-order one, in the injected FFN.Specifically, we first unfreeze the classifier layer of PLMs to make adapter-based fine-tuning. Next, we inject a one-layer FFN including a weight vector and a bias vector on the output of self-attention to update the network architecture.  After that, we load the trained parameters in the classifier layer and freeze other parameters except those in FFN and the followed normalization layer to make continuing fine-tuning. We carry out experiments on GLUE benchmark, including nine corpora. Experimental results demonstrate that our proposed kernel adapter is powerful which achieves competitive performance with extreme few parameters compared with full fine-tuning. It also has fewer parameters than any other previous adapters. 
%
Thus, a possible way to significantly reduce the scale of parameters for fine-tuning might be designing an adapter to work with the self-attention module in PLMs. We also find related research on adapter tuning that injects adapters into the self-attention layer~\citep{lyu2022study,lyu2023backdoor}, such as IA3~\citep{liu2022few}.
%
%The adapter takes the PLMs' multi-head self-attention output as its input and performs the transformation defined by the adapter module to output the results. Such transformed result is further used as the input of the next transformer layer.
%
There are three questions need to be answered when designing the adapter.
{\it {\bf Q1.} Where should the adapter that acts on self-attention outputs be injected into the PLMs?
{\bf Q2.} What is the suitable form of the adapter that satisfies both competitive performance and parameter-efficiency? 
{\bf Q3.} What other essential parameters should not be frozen in adapter tuning?}
To answer these questions, we conduct the following empirical studies: i) Analyzing the changes of self-attention outputs before and after full fine-tuning to verify the importance of self-attention which is therefore necessary to inject the adapter; ii) Comparing the difference among all fitting functions to select the suitable form of the adapter; iii) Analyzing the gradients of PLMs after fine-tuning on downstream tasks to select out the modules of great importance which should to be trained in the adapter tuning.
%
%Firstly, we compare the value of the self-attention outputs before and after the full fine-tuning process and analyze the changes on the output of self-attention caused by fine-tuning in order to verify the importance of self-attention.
%
%Next, we adopt different functions to fit self-attention outputs in full fine-tuning and adapter-based fine-tuning, respectively, and compare the difference among all fitting functions in order to select the suitable form of the adapter.
%
%Finally, we sort the gradients of PLMs after fine-tuning on downstream tasks in order to select out the layers of great importance for guiding adapter-based fine-tuning.

According to the empirical analysis, we propose a novel adapter tuning method as follows: We first learn the classification module to output prediction results on a given downstream task, without updating the PLMs' other parameters. Since the classification module is a linear model, this step requires light-weight computation cost.
Then we design an adapter and inject it right after the multi-head self-attention outputs of PLMs.
Particularly, we freeze all parameters except parameters in the designed adapter and the subsequent normalization module for continuous fine-tuning.
As there are usually multiple layers with the same architecture in PLMs, e.g., BERT~\citep{BERT} model of base version has 12 layers, we inject such an adapter module in each layer of PLMs. 
In designing the adapter, we only adopt element-wise linear transformation, rather than high-order ones, as the computational logic for the adapter.
Specifically, the adapter includes a weight vector and a bias vector which have the same dimension as the output of the multi-head self-attention module. The multi-head self-attention output is multiplied by the weight vector of the adapter using the element-wise product (also called the Hadamard product), then added by the corresponding bias vector to obtain new self-attention outputs.
Thus, we name the designed adapter as Hadamard adapter.

We carry out experiments on GLUE benchmark, including eight tasks. The experimental results demonstrate that the proposed Hadamard adapter achieves competitive performance with much fewer parameters than the existing fine-tuning methods.
In addition, we take the learned parameter values of the Hadamard adapter as representations of downstream tasks. 
Through further analysis, we summarize some valuable tuning patterns for Hadamard adapter shared by various downstream tasks, which provide valuable guidance for further parameter reduction using shared adapters in future research.

To summarize, our contributions in this paper are threefold:
% \vspace{-5pt}
\begin{itemize}
\setlength{\itemsep}{0pt}
    \item Based on comprehensive empirical analysis, we design Hadamard adapter, which acts on self-attention outputs in PLMs with element-wise linear transformation. We also design an extreme parameter-efficient adapter tuning method based on the Hadamard adapter.
    %which requires the fewest parameters compared to previous parameter-efficient adapters.
    
    \item We conduct extensive comparative experiments with several mainstream PLMs. The experimental results show that the proposed Hadamard adapter achieves the highest parametric efficiency in the fine-tuning history, and has competitive performance with full fine-tuning for various downstream tasks.

     \item We summarize some valuable tuning patterns for Hadamard adapter shared by various downstream tasks, which provide valuable guidance for further parameter reduction using shared adapters in future research.
     
    %The experimental results demonstrate the extreme parameter-efficiency of the proposed Hadamard adapter and its competitive performance compared with full fine-tuning in various downstream tasks.
\end{itemize}

%todo
\section{Empirical analysis}
%什么地方加。加成什么样。 2,2 2,1
To guide the design of our adapter tuning method, we conduct empirical studies that target at answering the three key questions as listed in the Introduction.
%
%Thus, a possible way to significantly reduce the scale of parameters for fine-tuning might be designing an adapter to work with the self-attention module in PLMs.
%
%The adapter takes the PLMs' multi-head self-attention output as its input and performs the transformation defined by the adapter module to output the results. Such transformed result is further used as the input of the next transformer layer.
%
%There are three questions need to be answered when designing the adapter.
%In this section, we introduce the empirical studies we conduct to verify the rationality and feasibility of our proposed parameter-efficient adapter tuning method.
% \yuyan{It aims to answer the following three questions: i) Why is the adapter injected after self-attention outputs? ii) What is the suitable form of the adapter that satisfies both competitive performance and parameter-efficiency? iii) What other necessary and important parameters should be unfrozen with adapter-based fine-tuning?}
%
In the following of this section, we first analyze the changes of self-attention outputs before and after full fine-tuning (for answering Q1), then we compare the difference among all fitting functions to select the suitable form of the Hadamard adapter (for answering Q2). Finally, we analyze the gradients of PLMs after fine-tuning on downstream tasks to select out the modules of great importance that would not be frozen in the adapter tuning (for answering Q3).

%\yuyan{Firstly, we compare the value of the self-attention outputs before and after the full fine-tuning process and analyze the changes of self-attention outputs caused by fine-tuning. In this way, we verify the importance of self-attention which is therefore necessary to inject the adapter after the outputs of it.
%
%Next, we adopt different functions to fit self-attention outputs in full fine-tuning and adapter tuning, respectively, and compare the difference among all fitting functions. In this way, we select the suitable form of the adapter both satisfying parameter-efficiency and competitive performance.
%
%Finally, we sort the gradients of PLMs after fine-tuning on downstream tasks. In this way, we select out the modules of great importance which should to be trained in the adapter tuning.}

\subsection{The changes of self-attention outputs}
\label{cop}
%To analyze the changes of the self-attention outputs before and after fine-tuning for PLMs, 

\begin{figure*}[t]
  \centering
  \includegraphics[width=0.82\linewidth]{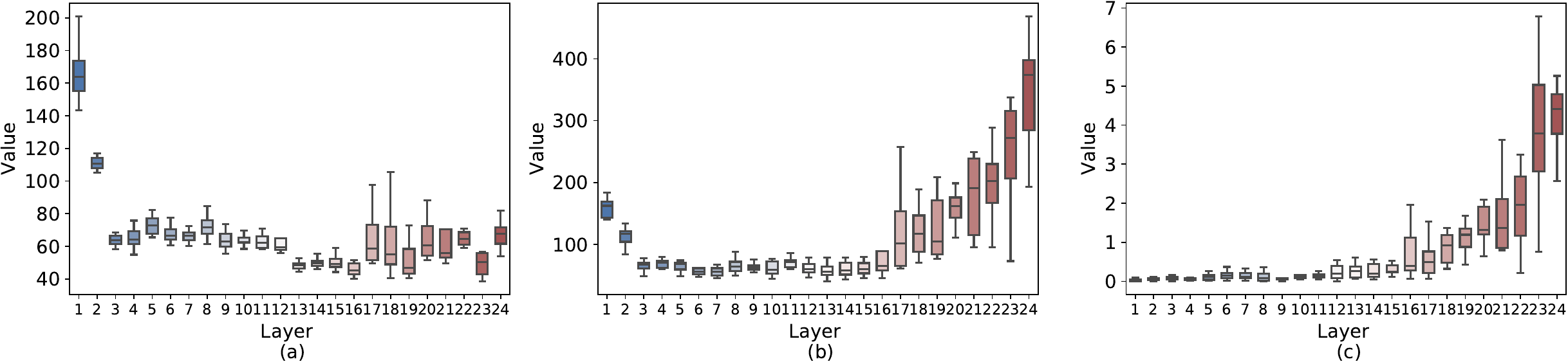}
  \caption{The distribution of the norm of the self-attention outputs among all tasks before (a) and after fine-tuning (b), and the corresponding changes (c) in each layer. 
  % In each layer, a box represents the distribution of characteristic values (a, b) or characteristic values changes (c) among all downstream tasks; line in the middle of a box represents the mean characteristic values (a, b) or changes (c) of all downstream tasks; the top and bottom of a box represent the upper and lower quartiles of the distribution of characteristic values (a, b) or changes (c), respectively, among all downstream tasks; and the upper and lower edges of a box represent the minimum and maximum characteristic values (a, b) or changes (c), respectively, among all downstream tasks.
  }
  \label{fig:changing}
\end{figure*}

We employ eight tasks in the GLUE benchmark to conduct the first analysis of how PLM's self-attention output changes before and after fine-tuning.
For PLM, we adopt Roberta-large model, which has 24 hidden layers and outputs 1024-dimensional tensors in the encoder, as an example to make analysis.
Specifically, in order to compare the changes of self-attention outputs in each layer among all tasks, we adopt the norm of self-attention outputs instead of the original self-attention outputs.
We analyze the distribution of the norm of self-attention outputs among all tasks before and after fine-tuning, and the changes during fine-tuning on each layer 
as shown in Fig.~\ref{fig:changing}.
%
% Specifically, we select nine tasks in the GLUE benchmark. 
The process is shown as the following equations:
\begin{gather}
\small
    % \overline{A_o}=\frac{1}{L}\frac{1}{H}\sum_{j=1}^L\sum_{i=1}^H a_{ij},\quad
    % \overline{A_o'}=\frac{1}{L}\frac{1}{H}\sum_{j=1}^L\sum_{i=1}^H a_{ij}',\quad
    \Vert \bm{A_b} \Vert_2=\sqrt{\lambda_{max}(\bm{A_b}^T\bm{A_b})},\quad
    \Vert \bm{A_a} \Vert_2=\sqrt{\lambda_{max}(\bm{A_a}^T\bm{A_a})}\\
    \Delta=\frac{\Vert \bm{A_a} \Vert_2-\Vert \bm{A_b} \Vert_2}{\Vert \bm{A_b} \Vert_2}
\end{gather}
where 
$\Vert \bm{A_b} \Vert_2$ and $\Vert \bm{A_b} \Vert_2$ represent
the norm of self-attention outputs among all tasks before and after fine-tuning in a hidden layer, and $\lambda_{max}(\bm{A_b}^T\bm{A_b})$ is the eigenvalue of the matrix $\Vert \bm{A_b} \Vert_2$.

One box in Fig.~\ref{fig:changing}(a)(b) and Fig.~\ref{fig:changing}(c) represents the distribution of the norm of self-attention outputs and the corresponding changes in the layer, respectively. 
As can be observed in Fig.~\ref{fig:changing}, the norm of self-attention outputs of all tasks significantly increase from an average of 60 to an average of 100 after fine-tuning, especially in the middle and back layer (Fig~\ref{fig:changing}(a)(b)). 
After the fifteen layers, the changes become more significant as the number of layers increases, reaching the greatest changes at the last layer (Fig~\ref{fig:changing}(c)). 
The above observations indicate that self-attention outputs change significantly during the fine-tuning process, which inspire us with the answer to Q1 as follows: \emph{It is proper to inject an adapter right after the self-attention outputs to achieve similar performance gains with fine-tuning while updating much fewer parameters.}

\subsection{Fitting full fine-tuning}
\begin{figure*}[!t]
  \centering
  \includegraphics[width=0.87\linewidth]{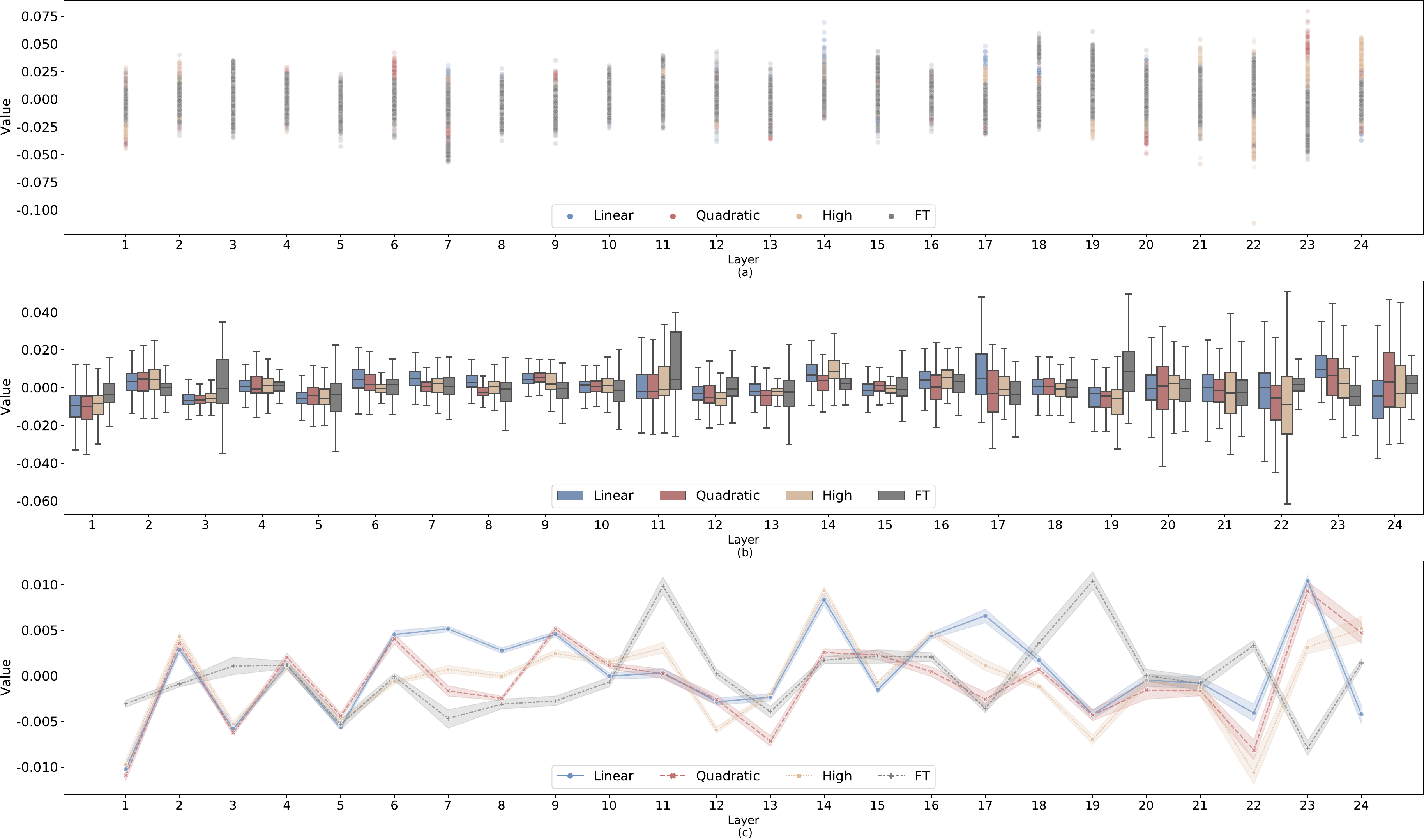}
  \caption{
  % The characteristic value distribution among all tasks (a), the average characteristic values of all tasks (b), and 
  The average values of each token in a sequence of all tasks (a), the characteristic value distribution among all tasks (b), and the average characteristic values of all tasks (c)
  % (a,b) and absolute value (c) of each sequence 
  % in a batch 
  % in the output of self-attention
  based on full fine-tuning and different fitting functions, respectively, in each hidden layer. 
  % In each layer, dots (a) of the same color represents the average value of each token in a sequence of all downstream tasks, and a box (b) represents the distribution of characteristic values among all downstream tasks, and a dot (c) represents the average characteristic values of all downstream tasks. 
  % Line in the middle of a box represents the mean characteristic values of all downstream tasks. The top and bottom of a box represent the upper and lower quartiles of the distribution of characteristic values, respectively, among all downstream tasks. And the upper and lower edges of a box represent the minimum and maximum characteristic values, respectively, among all downstream tasks.
  }
  \label{fig:fitting}
\end{figure*}

We design fitting functions for self-attention outputs to make adapter tuning, which aims at letting the values of self-attention outputs approximate those in full fine-tuning of PLMs. 
We first optimize the parameters in the classifier modules. Next, we reload them and train different fitting functions, including linear function, quadratic function and higher order function (i.e. cubic function), respectively, to obtain new self-attention outputs. 
%
% Inspired by the previous research~\citep{Parameter-Efficient_Transfer_Learning}, we also unfreeze and train the normalization modules together with the fitting functions. 
%
%c移到ab前面
After that, we calculate the average value of each token in a sequence through dividing by the hidden size of the PLM in the new self-attention outputs as shown in Fig.~\ref{fig:fitting}(a). The process is shown in the following equations:
\begin{gather}
\small
    a_j'=\frac{1}{H}\sum_{i=1}^H a_{ij}'
\end{gather} 
where $H$ represents the hidden size of a PLM. 
% $L$ represents the sequence length fed for the PLM. 
$a_{ij}'$ is the value of the $i^{th}$ dimension of the $j^{th}$ token in the new self-attention outputs in a hidden layer based on a task. $a_j'$ is the average value of each token in a sequence of a task.
More detailed, we then calculate the average value of each sequence through dividing by sequence length in the new self-attention outputs.
In this way, we obtain a characteristic value for each task which represents its respective average self-attention outputs. 
We analyze the distribution of the characteristic value among all tasks in each layer as shown in Fig.~\ref{fig:fitting}(b) with the process shown in the following equations:
\begin{gather}
\small
    a'=\frac{1}{L}\sum_{j=1}^L a_j'
\end{gather}
where 
% $H$ represents the hidden size of a PLM. 
$L$ represents the sequence length fed for the PLM. 
% $a_{ij}'$ is the value of the $i^{th}$ dimension of the $j^{th}$ token in the new self-attention outputs in a hidden layer based on a task. 
$a'$ is the characteristic value of a task.
We also analyze the average characteristic values of all tasks in each layer as shown in Fig.~\ref{fig:fitting}(c).

% More detailed, we also analyze the average values of each token in a sequence of all tasks

% average characteristic value of all tasks  in each layer as shown in Fig.~\ref{fig:fitting}(b) with the process shown in the following equations:
% \begin{gather}
% \small
%     \overline{A_o'}=\frac{1}{L}\frac{1}{H}\sum_{j=1}^L\sum_{i=1}^H a_{ij}'
% \end{gather}
% More detailed,

% and the average of the characteristic value among all tasks based on full fine-tuning and different fitting functions, respectively, and on each layer as shown in Fig.~\ref{fig:fitting}(a).
% The process is shown as the following equations:

% More detailed, we 

% In each layer, dots (a) of the same color represents the average value of each token in a sequence of all downstream tasks (see ), and a box (b) represents the distribution of characteristic values among all downstream tasks, and a dot (c) represents the average characteristic values of all downstream tasks.
As shown in Fig.~\ref{fig:fitting}(a), dots of the same color represents the average value of each token in a sequence of all downstream tasks corresponding to one of the three fitting functions and fine-tuning. The dots of four settings are covered by each other, which also proves that fitting functions of different orders are similar in approximating the performance of fine-tuning. 
As shown in Fig.~\ref{fig:fitting}(b), one box represents the distribution of characteristic values among all downstream tasks in a layer. Median, quartile ranges which correspond to the characteristic value distribution of different fitting functions and fine-tuning are similar in each hidden layer. 
When we analyze the average characteristic value of all tasks,
trends of linear function, quadratic function and higher order function are still similar, but slightly different from that of fine-tuning (Fig~\ref{fig:fitting}(c)). One dot represents the average characteristic values of all downstream tasks in a layer.
As the order increases, the values between the fitting function and fine-tuning are closer, but the difference in distance can be ignored compared with the increase in the number of parameters.
Therefore, we have answer to Q2 as follows: \emph{A linear function is qualified enough to act on self-attention outputs to fit the performance of fine-tuning.}

\subsection{Gradient Analysis}
\label{gra}
We output the gradient and unit gradient of the top five layers in the first and last epoch during training, respectively, of a PLM (such as BERT-base model). 
Two representative datasets MRPC (similarity and paraphrase task, 3.7k) and SST-2 (single-sentence classification, 67k) from the GLUE benchmark are selected for analysis, and the results are shown in Table ~\ref{tab:gradient}.
%
% Other tasks have similar results, as shown in Appendix~\ref{gra_more}.

\begin{table*}[t]
\small
\caption{\small The gradient and unit gradient of the top five layers (in descending order) in the first and last epoch, respectively. We adopt BERT-base model, MRPC and SST-2 dataset to show results and make analysis.}
    \begin{center}
        \begin{threeparttable}
        \resizebox{0.93\textwidth}{!}{
            \begin{tabular}{c|c|c||c|c}
                \toprule
                Task&\bf Gradient in first epoch&\bf Unit gradient in first epoch& \bf Gradient in last epoch &\bf Unit gradient in last epoch\\
                \midrule
                \makecell[l]{MRPC}
                &\tabincell{l}{\makecell[l]{classifier.weight} \\
                \makecell[l]{embeddings.token\_type\_embeddings.weight} \\
                \makecell[l]{encoder.layer.4.intermediate.dense.weight} \\
                \makecell[l]{encoder.layer.5.intermediate.dense.weight} \\
                \makecell[l]{encoder.layer.4.attention.self.value.weight}}
                &
                \tabincell{l}{\makecell[l]{classifier.bias} \\
                \makecell[l]{classifier.weight} \\
                \makecell[l]{embeddings.token\_type\_embeddings.weight} \\
                \makecell[l]{encoder.layer.4.output.LayerNorm.bias} \\
                \makecell[l]{encoder.layer.8.output.LayerNorm.weight}}
                &
                \tabincell{l}{\makecell[l]{classifier.weight} \\
                \makecell[l]{embeddings.token\_type\_embeddings.weight} \\
                \makecell[l]{pooler.dense.weight} \\
                \makecell[l]{encoder.layer.4.intermediate.dense.weight} \\
                \makecell[l]{encoder.layer.7.intermediate.dense.weight}}
                &
                \tabincell{l}{\makecell[l]{classifier.bias} \\
                \makecell[l]{classifier.weight} \\
                \makecell[l]{embeddings.token\_type\_embeddings.weight} \\
                \makecell[l]{encoder.layer.4.output.LayerNorm.bias} \\
                \makecell[l]{encoder.layer.3.output.LayerNorm.bias}}
                \\  
                \midrule
                \makecell[l]{SST-2} & \tabincell{l}{\makecell[l]{embeddings.token\_type\_embeddings.weight} \\
                \makecell[l]{embeddings.position\_embeddings.weight} \\
                \makecell[l]{embeddings.word\_embeddings.weight} \\
                \makecell[l]{encoder.layer.6.intermediate.dense.weight} \\
                \makecell[l]{encoder.layer.1.intermediate.dense.weight}}
                &
                \tabincell{l}{\makecell[l]{classifier.bias} \\
                \makecell[l]{embeddings.token\_type\_embeddings.weight} \\
                \makecell[l]{classifier.weight} \\
                \makecell[l]{embeddings.LayerNorm.bias} \\
                \makecell[l]{encoder.layer.3.output.LayerNorm.weight}}
                &
                % \\    
                % \midrule
                % \midrule
                % Task & \bf Gradient in last epoch &\bf Unit gradient in last epoch\\
                % \midrule
                % \makecell[l]{MRPC}
                % &
                % \tabincell{l}{\makecell[l]{classifier.weight} \\
                % \makecell[l]{embeddings.token\_type\_embeddings.weight} \\
                % \makecell[l]{pooler.dense.weight} \\
                % \makecell[l]{encoder.layer.4.intermediate.dense.weight} \\
                % \makecell[l]{encoder.layer.7.intermediate.dense.weight}}
                % &
                % \tabincell{l}{\makecell[l]{classifier.bias} \\
                % \makecell[l]{classifier.weight} \\
                % \makecell[l]{embeddings.token\_type\_embeddings.weight} \\
                % \makecell[l]{encoder.layer.4.output.LayerNorm.bias} \\
                % \makecell[l]{encoder.layer.3.output.LayerNorm.bias}}
                % \\  
                % \midrule
                % \makecell[l]{SST-2} &
                \tabincell{l}{\makecell[l]{classifier.weight} \\
                \makecell[l]{embeddings.token\_type\_embeddings.weight} \\
                \makecell[l]{embeddings.position\_embeddings.weight} \\
                \makecell[l]{embeddings.word\_embeddings.weight} \\
                \makecell[l]{encoder.layer.8.intermediate.dense.weight}}
                &
                \tabincell{l}{\makecell[l]{classifier.bias} \\
                \makecell[l]{classifier.weight} \\
                \makecell[l]{embeddings.token\_type\_embeddings.weight} \\
                \makecell[l]{encoder.layer.7.output.LayerNorm.weight} \\
                \makecell[l]{encoder.layer.7.output.LayerNorm.bias}}
                \\    
                \bottomrule
            \end{tabular}}
        \end{threeparttable}
    \end{center}
    \label{tab:gradient}
    % \vspace{-4mm}
\end{table*}

From the results of the gradients, we speculate that the classifier weights, embedding weights and intermediate weights of all tasks are more important in fine-tuning than the other modules because their gradients contribute the most in the first and last epochs.
However, we find that some modules which contribute most during the fine-tuning have a large number of parameters, such as the intermediate module, which accounted for nearly half of all parameters.
If we still fine-tune them, a lot of redundant computation is inevitable.
We further analyze the results of unit gradient, which is the gradient divided by the number of parameters, and the results show that the classifier weights, embedded weights, and normalized weights are more important because their unit gradients contribute the most in the first and last epochs.

Therefore, we have answer to Q3 as follows: \emph{We select out the classifier and the normalization as trainable modules in the adapter tuning.} We also make a theoretical analysis of the selection of these parameters as follows: 
i) The classifier makes prediction in the downstream tasks and directly affects the performance of tasks.
ii) Normalization is to limit the data within a certain range and unify the distribution of each batch of training data.
Since the input data distribution of each batch of the network is constantly changing, the changes of training pattern without normalization will make it difficult for the network to find a balance point, thus affecting the convergence of the network.

%Therefore, we have answer to Q3 as follows:

%\noindent{\bf Answer to Q3.} 

\section{Methodology}
In this section, we first introduce the details of Hadamard adapter, and then clarify the process of the proposed parameter-efficient adapter tuning method in solving downstream tasks. 
% Specifically, we first only unfreeze the pooling and classifier layers to train the classifier. Next, we inject a linear feedforward network (FFN) including weight vectors and bias vectors which works on the output of self-attention. After that, we reload the trained pooling and classifier layers, and only retrain the linear FFN and the normalization layers.
%The overall framework is shown in Fig~\ref{fig:framework}.

\begin{figure}[!t]
  \centering
  \includegraphics[width=0.88\linewidth]{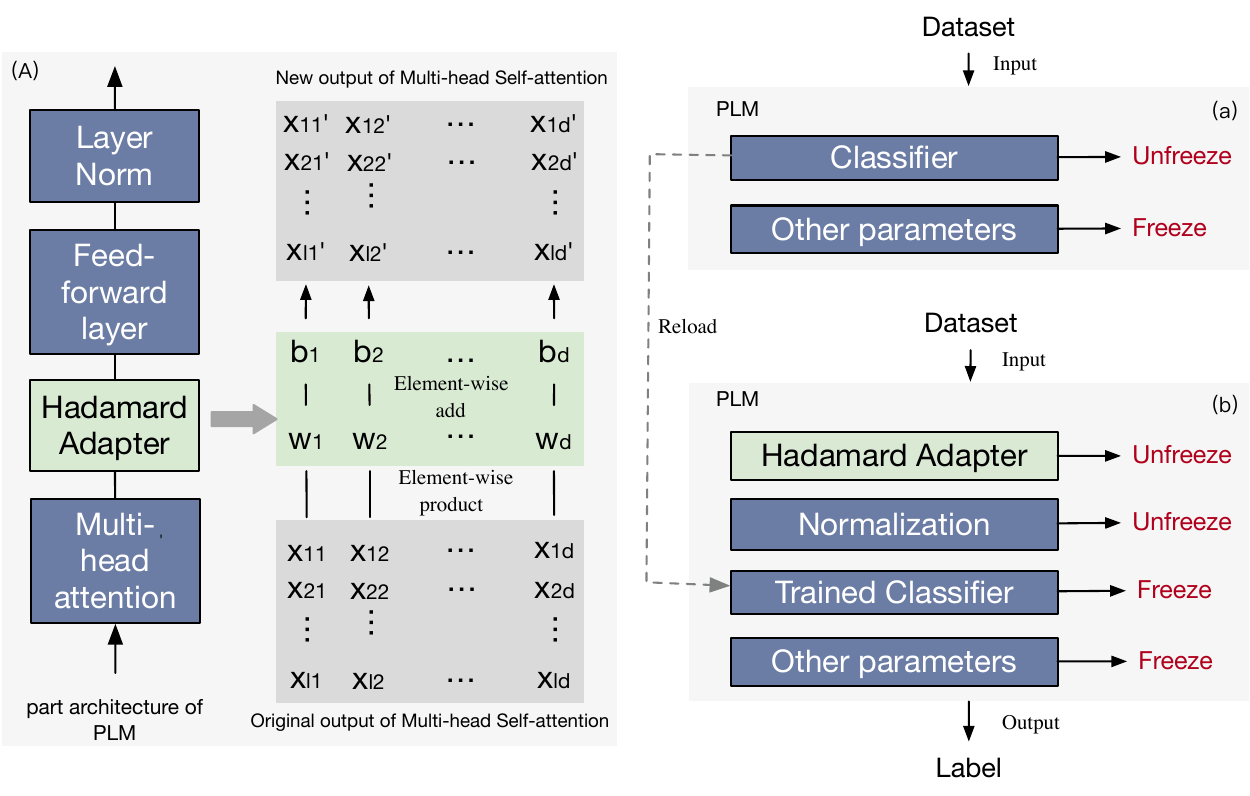}
  \caption{The framework of the Hadamard adapter (A), and the process of the parameter-efficient adapter tuning method, including two parts: (a) Train the classifier; (b) Inject the Hadamard adapter for self-attention outputs and unfreeze the normalization module.}
  \label{fig:framework}
\end{figure}

\subsection{The Hadamard adapter}
The framework of the Hadamard adapter is shown in Fig.~\ref{fig:framework}(A). It is equivalent to a linear transformation based on feature dimension as shown in the following equation: 
\begin{gather}
\small
    Adap: \bm{A_{ij}'}=\bm{W_j}*\bm{A_{ij}}+\bm{b_j}
\end{gather}
Different feature dimensions correspond to different linear transformation parameters, but different positions (i.e. each token in a sequence) share the same parameters. We put such an adapter in each layer of the PLMs. Adapters of different layers have different parameters. Based on the empirical analysis in Sec.~\ref{cop}, we put this adapter right after the self-attention outputs. 
Specifically, we multiply the weight vector $\bm{W_j}$ of the Hadamard adapter to the self-attention outputs $\bm{A_{ij}}$, and then add the bias vector $\bm{b_j}$ to the product to obtain new self-attention outputs $\bm{A_{ij}'}$. The weight vector and the bias vector in the Hadamard adapter are both 1-d vectors of the same shape as the hidden size of the PLM, such as 768 for the base version or 1024 for the large version for BERT model. The number of the Hadamard adapter in a PLM is also the same as the number of PLMs' layers, such as 12 for the base version and 24 for the large version in BERT model. All weight vectors are initialized as 1.0 and all bias vectors are initialized as 0.0. The initial value is equivalent to not adding any adapter.
The approximate number of parameters of this adapter is from 30,000 to 100,000 according to different size of PLMs and 3,000 to 4,000 in each layer of PLMs, which is only 0.03\% of the full fine-tuning.

\subsection{The adapter tuning method}
The parameter-efficient adapter tuning method is to inject adapters to PLMs and make continuous fine-tuning as shown in Fig.~\ref{fig:framework}(a)(b).
Specifically, we first only unfreeze and train the pooling and classifier modules on downstream tasks. Next, we inject the Hadamard adapter right after self-attention outputs. After that, we reload the trained pooling and classifier layers, and only fine-tune the Hadamard adapter and the normalization module in each layer. 
Although the above-mentioned two-stage training process can be time-consuming and computationally expensive, the performance is better than joint training in our experiments. 
We have analyze the possible reason in the paper that the proposed Hadamard Adapter has not been optimized in the pre-training stage which will affect the effect of classifier layer. We also find other research, such as LP-FT~\citep{kumar2022fine}, demonstrating the better performance of two-stage training.

%\subsection{Train the classifier}
\textbf{Train the classifier module. }
For each given downstream task, we first learn only the classifier module on the training dataset, including the pooling and linear output layers. While learning the classifier module, the other parameters of the PLMs are frozen.
Since all layers except the classifier module do not participate in backward propagation and gradient update, they can be shared by different tasks. In addition, the computational cost of this step is very small compared to the overall cost of full fine-tuning.
%We first only unfreeze the pooling and the classifier layers to fine-tune the classifier for different tasks. Other layers will be freezed when fine-tuning the classifier. It represents that layers except the pooling and the classifier layers will not attend backward propagation and gradient update, and therefore may be shared by many tasks.
%
%As the layers except the final classification module will not attend backward propagation and gradient update, therefore they can be shared by different tasks. Also, the computational cost of this step is very small compared with the whole cost of the standard fine-tuning. 

% Because the final classification layer is very important for prediction, in order to reduce the disturbance of the inserted norm layer or incomplete parameters compared with full fine-tuning, we only fine-tune the pooling and the classifier layers and save the fine-tuned parameters.

\textbf{Inject the Hadamard adapter and unfreeze the normalization module. }
Secondly, we inject the Hadamard adapter right after the self-attention outputs. Multi-head self-attention is an important mechanism in PLMs, which first obtains representation of queries $Q$, keys $K$ and values $V$, and then makes scaled dot-product for $heads=16$ times to obtain self-attention score of each attention head. 
Outputs of all attention heads are finally concatenated to obtain self-attention output $A_i$.
Based on multi-head self-attention, we first transform the 3-D dimensional self-attention outputs into 2-D dimensional matrices, that is, making the first and second dimension flatten. 
Next, we input the 2-D dimensional self-attention outputs into the Hadamard adapter and transform the self-attention outputs into 3-D dimensional matrix $A_i'$. The process is as follows:
% Specifically, we multiply the weight vector $\bm{W^N}$ to the 2-D dimensional attention output, and add the bias vector $\bm{b^N}$ to the product to obtain new attention output $A_i^N$. The weight vector and the bias vector in the linear FFN are both 1-d vectors of same shape as the hidden size of the PLM, such as 768 for base version or 1024 for large version. The number of the linear FFN are also the same as the PLMs' layer, such as 12 layers for the base version or 24 layers for large version. All weight vectors are initialized as 1 and all bias vectors are initialized as 0.
\begin{gather}
\footnotesize
    Q_l^i=Q\bm{W_l^Q},K_t^i=K\bm{W_l^K},V_t^i=V\bm{W_l^V},\quad
    A_l^i=softmax(\frac{{Q_l^iK_l^i}^T}{\sqrt{d_k}})V_l^i\\
    A_i=Concat(A_1^i,\cdots,A_T^i),\quad
    A_i'=Adap(A_i)
\end{gather}
%
% \textbf{Unfreeze the normalization modules. }
After that, We reload the parameters of the trained classifier module and only unfreeze the normalization module besides the Hadamard adapter. 
Based on the empirical analysis in Sec.~\ref{gra}, we argue that the normalization module is very important to improve the accuracy of downstream tasks. 
The reason is that the value range of the self-attention outputs changes after being transformed with the Hadamard adapter. 
Obviously, it is necessary to relearn the parameters of the normalized module and make them adapt to the value distribution of the new self-attention outputs in order to achieve better results.
%Then it's obvious that the parameters of normalization modules should be relearned to adapt themselves to the value distribution of the new self-attention output to achieve good results.
% we consider the normalization layers can speed up the coverage, which is more important in the adapter-tuning setting.
In order to further scale the parameter size and reach the limit of parameter efficiency, we only unfreeze the normalization module right after the intermediate outputs rather than unfreeze that right after the self-attention output.
%
%In order to further scale the parameter size to achieve extreme parameter efficient, we only unfreeze the normalization modules in each layer instead of those right after the self-attention outputs.
% The trainable parameters for PLMs of base and large version are both \texttt{0.03\%} of the corresponding full fine-tuning. 

\section{Experiment}
This section reports the experiments conducted on GLUE benchmark in validating the effectiveness of the proposed Hadamard adapter on different SOTA PLMs, as well as comparing the performance with other parameter-efficient methods. 
%
%We make comprehensive analysis for the pattern of the proposed parameter-efficient adapter-tuning method among different downstream tasks in each layer. 
%
% Then we conduct ablation study to investigate the function of each module in the proposed adapter-tuning method.

\subsection{Experimental setup}
The experiments are carried out on Tesla V100 GPUs with Pytorch in Python. Similar with previous work~\citep{BERT,RoBERTa,BART,DeBERTa,ELECTRA}, the batch size is set to 16 or 32, and the sequence length is set to 128. We maintain the hyper parameters such as 0.01 for weight decay, 0.9 for $\beta_1$, and 0.999 for $\beta_2$, etc. We set the learning rate from 2e-03 to 4e-03 in training the classifier module, from 2e-05 to 4e-05 in full fine-tuning, and from 1e-03 to 9e-03 in training with the Hadamard adapter to obtain the best performance in each training setting. All tasks are training for 20 epochs with each PLM.

\textbf{Datasets and Baselines. }
We adopt GLUE benchmark~\citep{GLUE}, including eight tasks from different domains, as the datasets. They are divided into single-sentence classification (CoLA, SST-2), similarity and paraphrase (MRPC, STS-B, QQP), and inference (MNLI, QNLI, RTE). We ignore the WNLI dataset whose data amount is too small. Same as the previous research~\citep{BitFit,LORA} on GLUE benchmark, we also adopt Matthews correlation coefficient~\footnote{https://en.wikipedia.org/wiki/Phi\_coefficient} and Pearson coefficient~\footnote{https://en.wikipedia.org/wiki/Pearson\_correlation\_coefficient} to evaluate the performance on the CoLA dataset and STS-B dataset, respectively, and use accuracy for other datasets. We carry out our experiments on several SOTA PLMs, including BERT~\citep{BERT}, RoBERTa~\citep{RoBERTa}, BART~\citep{BART}, DeBERTa~\citep{DeBERTa}, and ELECTRA~\citep{ELECTRA}. Moreover, we use powerful parameter-efficient adapters shown in Table~\ref{tab:result1} as baselines to compare the parameter amount and performance on GLUE benchmark.

\subsection{Main results}
We first report the performance of the classifier, Hadamard adapter and full fine-tuning in GLUE benchmark based on several SOTA PLMs in Table~\ref{tab:result}, where the results all come from our own machines, which represent only fine-tuning the classifier module, the proposed adapter tuning method, and full fine-tuning, respectively. 
We find that only training the classifier module achieves 77.5\% in performance compared with that of the corresponding full fine-tuning averagely in selected PLMs and tasks.
It represents that the classifier module is important for solving downstream tasks, so it is necessary to be fine-tuned alone.
Next, we reload the trained classifier module, inject the Hadamard adapter for self-attention outputs, and only unfreeze the normalization module. The performance increase a large degree, which achieves 99.4\% compared with that of full fine-tuning averagely in selected PLMs and tasks. Some results are even better than the corresponding full fine-tuning, such as the MPRC dataset with the BERT-base model. It represent that the Hadamard adapter has a stunning effect which has competitive performance and achieve extreme parameter efficiency (0.033\% parameters of full fine-tuning).

\begin{table*}[t]
\footnotesize
\caption{\footnotesize Performance of training classifier module, adapter tuning and full fine-tuning in GLUE benchmark based on several SOTA PLMs.}
    \begin{center}
        \begin{threeparttable}
        \resizebox{0.7\textwidth}{!}{
            \begin{tabular}{c|c|ccccccccc}
                \toprule
                % \multirow{2}{*}{\bf PLMs} &\multirow{2}{*}{\bf Training type}& \multicolumn{9}{c}{\bf Downstream tasks} \\
                \bf PLMs&\bf Training type
                % &\bf Trainable parameters
                &\bf MRPC&\bf CoLA&\bf MNLI&\bf QNLI&\bf QQP&\bf RTE&\bf SST-2&\bf STS-B&\bf Average\\	
                \midrule
                \multirow{3}{*}{\bf BERT-base}
                &Classifier
                % &-
&71.8
&37.0
&54.4
&70.6
&79.3
&57.4
&87.4
&60.3&64.8\\
                &Hadamard adapter
                % &0.04M
&\bf 90.2 
&\bf 58.4 
&\bf 80.4 
&\bf 89.7 
&\bf 85.9 
&\bf 71.9 
&\bf 92.4 
&\bf 88.5&\bf 82.2\\
                &Full fine-tuning
                % &110M
&89.4
&56.5
&83.9
&91.3
&87.5
&64.6
&93.0
&88.6&81.9\\
                \midrule
                \multirow{3}{*}{\bf BERT-large}
                &Classifier
                % &-
&72.2
&37.9
&61.7
&71.8
&79.8
&58.1
&89.7
&61.9&66.6\\
                &Hadamard adapter
                % &
&\bf 91.3 
&\bf 62.7 
&\bf 83.6 
&\bf 91.1 
&\bf 87.2 
&\bf 73.1 
&\bf 93.1 
&\bf 90.0&\bf 84.0\\
                &Full fine-tuning
                % &335M
&90.2
&62.8
&85.6
&92.1
&88.5
&71.9
&93.1
&90.1&84.3\\
                \midrule
                \multirow{3}{*}{\bf RoBERTa-base}
                &Classifier
                % &-
&73.1
&44.4
&59.3
&70.9
&75.4
&58.5
&83.6
&61.9&65.9\\
                &Hadamard adapter
                % &
&\bf 92.1 
&\bf 64.3 
&\bf 85.5 
&\bf 91.6 
&\bf 87.7 
&\bf 80.7 
&\bf 93.9 
&\bf 90.1&\bf 85.7\\
                &Full fine-tuning
                % &125M
&91.9
&64.4
&86.7
&92.8
&89.0
&78.0
&94.7
&91.1&86.1\\
                \midrule
                \multirow{3}{*}{\bf RoBERTa-large}
                &Classifier
                % &-
&72.8
&45.9
&62.0
&70.6
&77.3
&58.8
&88.4
&62.4&67.3\\
                &Hadamard adapter
                % &
&\bf 92.5 
&\bf 67.8 
&\bf 89.5 
&\bf 95.1 
&\bf 90.3 
&\bf 85.9 
&\bf 96.1 
&\bf 91.9&\bf 88.6\\
                &Full fine-tuning
                % &355M
&92.7
&65.9
&91.0
&94.5
&89.7
&86.3
&96.2
&91.7&88.5\\
                \midrule
                \multirow{3}{*}{\bf BART-base}
                &Classifier
                % &-
&71.5
&43.7
&60.1
&70.0
&76.1
&56.9
&87.3
&61.2&65.9\\
                &Hadamard adapter
                % &
&\bf 86.2
&\bf 61.7
&\bf 85.5
&\bf 92.8
&\bf 86.8
&\bf 76.5
&\bf 94.1
&\bf 89.4&\bf 84.1\\
                &Full fine-tuning
                % &140M
&87.1
&62.0
&87.8
&93.9
&88.3
&77.7
&95.0
&91.2&85.4\\
                \midrule
                \multirow{3}{*}{\bf BART-large}
                &Classifier
                % &-
&72.3
&44.3
&61.8
&71.4
&78.0
&58.4
&88.9
&61.8&67.1\\
                &Hadamard adapter
                % &
&\bf 88.1 
&\bf 65.0 
&\bf 87.4 
&\bf 94.0 
&\bf 89.0 
&\bf 82.3 
&\bf 95.7
&\bf 91.3&\bf 86.6\\
                &Full fine-tuning
                % &400M
&88.2
&63.4
&88.0
&95.5
&90.1
&79.3
&95.9
&91.9&86.5\\
                \midrule
                \multirow{3}{*}{\bf DeBERTa-base}
                &Classifier
                % &-
&72.2
&45.2
&62.1
&70.3
&76.6
&58.8
&87.8
&63.5&67.1\\
                &Hadamard adapter
                % &
&\bf 89.0
&\bf 65.8
&\bf 90.1
&\bf 94.7
&\bf 89.2
&\bf 85.4
&\bf 94.4
&\bf 92.4&\bf 87.6\\
                &Full fine-tuning
                % &134M
&91.3
&66.2
&91.2
&96.1
&90.2
&87.4
&96.1
&93.0&88.9\\
                \midrule
                \multirow{3}{*}{\bf DeBERTa-large}
                &Classifier
                % &-
&73.5
&46.3
&63.9
&72.3
&77.9
&60.2
&89.1
&64.1&68.4\\
                &Hadamard adapter
                % &
&\bf 90.4
&\bf 67.3
&\bf 92.4
&\bf 95.9
&\bf 89.8
&\bf 86.7
&\bf 96.8
&\bf 93.0&\bf 89.0\\
                &Full fine-tuning
                % &390M
&92.5
&68.0
&93.3
&96.9
&90.5
&88.3
&97.1
&93.5&90.0\\
                \midrule
                \multirow{3}{*}{\bf ELECTRA-base}
                &Classifier
                % &-
&72.9
&47.3
&63.3
&71.3
&76.6
&60.2
&89.5
&65.0&68.3\\
                &Hadamard adapter
                % &
&\bf 89.4
&\bf 68.2
&\bf 91.2
&\bf 96.1
&\bf 87.7
&\bf 85.6
&\bf 96.2
&\bf 93.5&\bf 88.5\\
                &Full fine-tuning
                % &110M
&90.1
&69.0
&92.9
&96.4
&88.5
&86.3
&96.9
&93.7&89.2\\
                \midrule
                \multirow{3}{*}{\bf ELECTRA-large}
                &Classifier
                % &-
&74.0
&47.1
&64.8
&72.9
&78.3
&61.2
&90.3
&66.8&69.4\\
                &Hadamard adapter
                % &
&\bf 92.0
&\bf 68.5
&\bf 92.9
&\bf 96.1
&\bf 89.9
&\bf 87.2
&\bf 97.1
&\bf 94.2&\bf 89.7\\
                &Full fine-tuning
                % &335M
&93.0
&69.3
&93.9
&96.4
&91.4
&88.7
&97.1
&94.9&90.6\\
                \bottomrule
            \end{tabular}}
        \end{threeparttable}
    \end{center}
    \label{tab:result}
    % \vspace{-4mm}
\end{table*}

\begin{table*}[t]
\footnotesize
\caption{\footnotesize Comparison between tuning with Hadamard adapter and other adapters in GLUE benchmark in several SOTA PLMs.}
    \begin{center}
        \begin{threeparttable}
        \resizebox{0.85\textwidth}{!}{
            \begin{tabular}{c|c|c|cccccccccc}
                \toprule
                % \multirow{2}{*}{\bf PLMs} &\multirow{2}{*}{\bf Adapter}& \multirow{2}{*}{\bf Parameters}& \multicolumn{10}{c}{\bf Downstream tasks} \\
                \bf PLMs&\bf Adapter& \bf Parameters&\bf MRPC&\bf CoLA&\bf MNLI&\bf QNLI&\bf QQP&\bf RTE&\bf SST-2&\bf STS-B&\bf Average\\	
                \midrule
                \multirow{2}{*}{\bf BERT-base}
                &Hadamard adapter&\bf0.03\%($\downarrow$ 0.06\%)& 90.2& 58.4&80.4&89.7&\bf  85.9& 71.9& \bf 92.4& 88.5&\underline{82.2}($\downarrow$ 0.2)\\
                &BitFit~\citep{BitFit}&\underline{0.09\%}& \bf90.4&\bf58.8&\bf81.8&\bf90.2&84.0& \bf72.3&92.1&\bf89.2&\bf 82.4\\
                \midrule
                \multirow{4}{*}{\bf BERT-large}
                &Hadamard adapter&\bf 0.03\%($\downarrow$ 0.05\%)&91.3& 62.7&83.6& 91.1&\bf 87.2&  73.1& 93.1& 90.0&\underline{84.0}($\downarrow$ 0.2)\\
                &BitFit&\underline{0.08\%}& \bf91.7&\bf63.6&84.6&\bf91.4&85.4& \bf73.2&93.2&\bf90.3&\bf 84.2\\
                &Adapters (8-256)~\citep{Parameter-Efficient_Transfer_Learning}&14.44\%& 89.5&59.5&\bf 85.0&90.7& 71.8&71.5 &94.0& 86.9&80.0\\%
                &Adapters (64)~\citep{Parameter-Efficient_Transfer_Learning}&13.33\%&  89.6&56.9&\bf 85.0& \bf 91.4&71.8& 68.8&\bf94.2& 87.3&79.6\\%Parameter-Efficient Transfer Learning for NLP
                % &CHILD-TUNINGF&& 91.22 (91.85)&63.71 (66.06)&&&&72.06 (74.73)&&&90.18 (90.92)&79.29\\
                % &CHILD-TUNINGD && 91.42 (92.17)&64.92 (66.03)&&&& 73.14 (76.17)&&& 90.18 (90.64)&79.92\\
                % &CHILD-TUNINGD + R3F&&92.23 (92.65)&65.18 (66.03)&&&&73.43 (76.17)&&& 90.18 (90.64)&80.26\\%Raise a Child in Large Language Model:Towards Effective and Generalizable Fine-tuning
                \midrule
                \multirow{5}{*}{\bf RoBERTa-base}
                &Hadamard adapter&\bf 0.03\%($\downarrow$ 0.21\%)&92.1&\bf 64.3& 85.5& 91.6 &87.7 &80.7 &93.9&90.1&\underline{85.7}($\downarrow$ 1.5)\\
                &BitFit&0.08\%&\bf92.7&62.0&84.7&91.8&84.0&81.5&93.7&90.8&85.2\\%BitFit: Simple Parameter-efficient Fine-tuning for Transformer-based Masked Language-models
                &Adpt$^D$~\citep{Adapterdrop}&0.24\%&88.5&60.8&87.1&93.1&90.2&71.5&94.2&89.7&84.4\\%LORA: LOW-RANK ADAPTATION OF LARGE LANGUAGE MODELS
                &Adpt$^D$&0.72\%&88.4&62.6&87.3&93.0&90.6&75.9&94.7&90.3&85.4\\%LORA: LOW-RANK ADAPTATION OF LARGE LANGUAGE MODELS
                &LoRA~\citep{LORA}&\underline{0.24\%}&89.7&63.4&\bf87.5&\bf93.3&\bf90.8&\bf86.6&\bf95.1&\bf91.5&\bf 87.2\\%LORA: LOW-RANK ADAPTATION OF LARGE LANGUAGE MODELS
                \midrule
                \multirow{9}{*}{\bf RoBERTa-large}
                &Hadamard adapter&\bf 0.03\%($\downarrow$ 0.2\%)&92.5& 67.8& 89.5& \bf 95.1 & 90.3& 85.9&  96.1& 91.9&\underline{88.6}(-)\\
                &Adpt$^P$~\citep{AdapterFusion}&0.85\%&90.2&\bf 68.3&90.2&94.8&91.9&83.8&96.1&92.1&88.4\\%LORA: LOW-RANK ADAPTATION OF LARGE LANGUAGE MODELS
                &Adpt$^P$&0.23\%&89.7&67.8&90.5&94.8&91.7&80.1&\bf96.6&91.9&87.9\\%LORA: LOW-RANK ADAPTATION OF LARGE LANGUAGE MODELS
                &Adpt$^H$~\citep{Parameter-Efficient_Transfer_Learning}&1.70\%&88.7&66.5&89.9&94.7&\bf 92.1&83.4&89.9&91.0&87.8\\%LORA: LOW-RANK ADAPTATION OF LARGE LANGUAGE MODELS
                &Adpt$^H$&0.23\%&87.7&66.3&90.3&94.7&91.5&72.9&90.3&91.5&86.4\\%LORA: LOW-RANK ADAPTATION OF LARGE LANGUAGE MODELS
                &LoRA&\underline{0.23\%}&90.2& 68.2&\bf 90.6&94.8&91.6&85.2&90.6&\bf92.3&\bf88.6\\%LORA: LOW-RANK ADAPTATION OF LARGE LANGUAGE MODELS
                
                &RoBERTa-AT~\citep{Y-Tuning}&0.85\%&\bf 92.9& 67.4&90.4& 94.7& 88.5&83.4& 96.3&-&87.7\\%Y-Tuning: An Efficient Tuning Paradigm for Large-Scale Pre-Trained Models via Label Representation Learning

                &RoBERTa-WARP~\citep{Y-Tuning}&0.28\%&91.2& 60.6& 88.2&  93.5& 84.5& \bf 86.3& 96.0&-&85.8\\%Y-Tuning: An Efficient Tuning Paradigm for Large-Scale Pre-Trained Models via Label Representation Learning

                &RoBERTa-YT~\citep{Y-Tuning}&4.60\%& 85.0& 54.4&83.1&  88.2&  87.4& 81.9& 94.5&-&82.1\\%Y-Tuning: An Efficient Tuning Paradigm for Large-Scale Pre-Trained Models via Label Representation Learning
                \midrule
                \multirow{3}{*}{\bf BART-large}
                &Hadamard adapter&\bf 0.02\%($\downarrow$ 7.71\%)&\bf 88.1& \bf 65.0 &\bf 87.4& \bf 94.0& \bf 89.0& \bf 82.3& \bf 95.7&-&\underline{86.6}($\uparrow$ 9.7)\\
                &BARTen-FbT~\citep{Y-Tuning}&8.52\%&76.0 &42.1&81.9&88.4&86.7&60.6& 93.2&-&75.6\\%Y-Tuning: An Efficient Tuning Paradigm for Large-Scale Pre-Trained Models via Label Representation Learning
                &BARTen-YT~\citep{Y-Tuning}&\underline{7.73\%}&79.2&44.4&82.3&88.2&85.5&62.8& 94.4&-&\bf 76.9\\%Y-Tuning: An Efficient Tuning Paradigm for Large-Scale Pre-Trained Models via Label Representation Learning
                \bottomrule
            \end{tabular}}
        \end{threeparttable}
    \end{center}
    \label{tab:result1}
    % \vspace{-4mm}
\end{table*}

We also compare the performance between the proposed adapter tuning method with Hadamard adapter and other parameter-efficient baselines as shown in Table~\ref{tab:result1}, where baselines' results are replicated from their corresponding published papers.
As can be observed, there is not much difference in performance, but the Hadamard adapter uses the fewest amount of parameters. 
This comparison results prove that the Hadamard adapter could uses the fewest parameters to achieve competitive performance with full fine-tuning and the existing parameter-efficient tuning methods.

%which is the most parameter-efficient algorithm in solving NLP-related downstream tasks.

\subsection{Ablation study}
\begin{figure}[!t]
  \centering
  \includegraphics[width=0.92\linewidth]{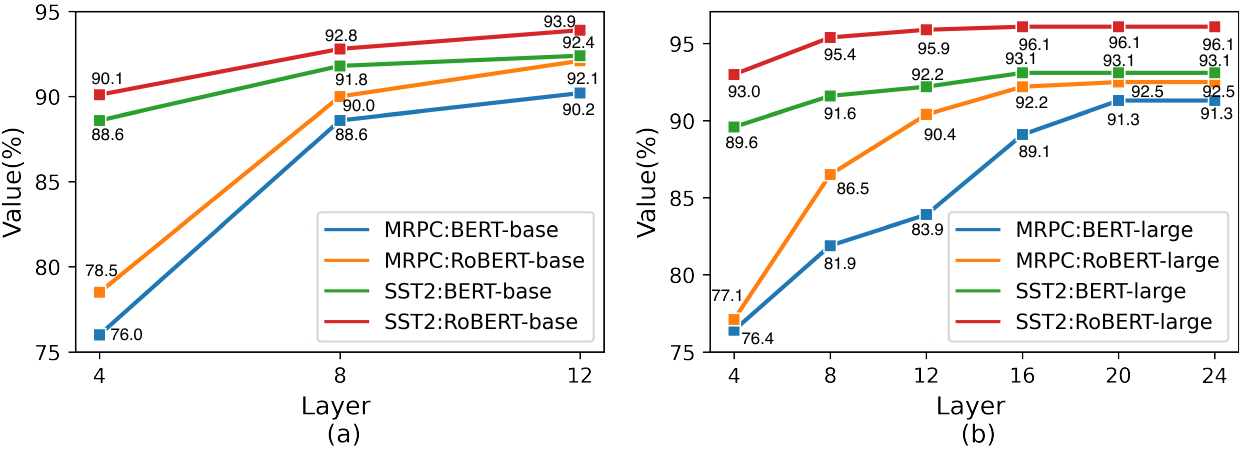}
  \caption{The influence of different number of unfreezing layers of the Hadamard adapter on the performance of the Hadamard adapter with model of base version (a) and large version (b).}
  \label{fig:layer}
\end{figure}

In this part, we carry out ablation study to clarify the effect of each module in the Hadamard adapter as shown in Table~\ref{tab:com}, as well as the influence of different number of unfreezing layers to the performance as shown in Fig.~\ref{fig:layer}.  
% Due to space limitation, we select several representative PLMs and downstream tasks to make analysis. 

\begin{table}[!t]
\footnotesize
\caption{\footnotesize The effect of each module in the Hadamard adapter on other downstream tasks based on BERT-base model. W: Weight, B: Bias, N: Norm, A: Att-Norm.}
    \begin{center}
        \begin{threeparttable}
        \resizebox{0.45\textwidth}{!}{
            \begin{tabular}{l|cccccccc}
                \toprule
               % \multirow{2}{*}{\bf Components} & \multicolumn{2}{c}{\bf Downstream tasks} \\
                \bf Module&\bf MRPC&\bf SST-2&\bf CoLA&\bf QNLI&\bf QQP&\bf MNLI&\bf RTE&\bf STS-B\\
                \midrule
                W&73.1&88.7&55.0&84.8&82.4&77.6&67.3&84.7\\
                B&81.0&91.1&56.9&88.0&84.9&79.6&68.7&86.0\\
                N&80.2&90.8&56.6&87.5&84.4&79.3&68.4&85.7\\
                A&79.6&90.5&56.5&87.3&84.1&79.0&68.2&85.5\\
                \midrule
                W+A&80.9&91.0&56.8&88.1&84.7&79.5&68.8&86.2\\
                W+N&81.9&91.3&57.2&88.3&85.0&79.6&69.0&86.3\\
                B+A&81.7&91.4&57.1&88.4&85.0&79.7&69.1&86.5\\
                B+N&82.1&91.7&57.2&88.6&85.2&79.8&69.3&86.7\\
                W+B&78.2&90.0&56.0&86.8&83.6&78.5&67.8&85.0\\
                \midrule
                W+B+N+A&82.6&92.0&57.3&88.9&85.4&80.0&69.8&87.2\\
                W+B+A&82.8&92.1&57.8&89.1&85.2&79.8&70.0&87.9\\
                \midrule
               \bf (Ours)&\bf 83.7&\bf 92.4&\bf 58.4&\bf 89.7&\bf 85.9&\bf 80.4&\bf 71.9&\bf 88.5\\
                \bottomrule
            \end{tabular}}
        \end{threeparttable}
    \end{center}
    \label{tab:com}
    % \vspace{-4mm}
\end{table}

\begin{table*}[!t]
\footnotesize
\caption{\footnotesize The influence of different number of unfreezing layers of the Hadamard adapter on the performance of downstream tasks.}
    \begin{center}
        \begin{threeparttable}
        \resizebox{0.72\textwidth}{!}{
            \begin{tabular}{c|c|cccccc||c|c|cccccc}
                \toprule
                % \multirow{2}{*}{\bf PLMs} &\multirow{2}{*}{\bf Downstream tasks& \multicolumn{6}{c}{\bf Unfreezing layers} \\
                \bf PLMs&\bf Tasks&\bf 4&\bf 8&\bf 12&\bf 16&\bf 20&\bf 24&\bf PLMs&\bf Tasks&\bf 4&\bf 8&\bf 12&\bf 16&\bf 20&\bf 24\\	
                \midrule
                \multirow{5}{*}{\bf BERT-base}
                                &CoLA&50.0
&56.3
&\bf 58.4
&-
&-
&-&\multirow{5}{*}{\bf RoBERTa-base}
                                &CoLA&55.7
&62.6
&\bf 64.3
&-
&-
&-\\
                &QNLI&80.8
&87.5
&\bf 89.7
&-
&-
&-&&QNLI&83.3
&89.2
&\bf 91.6
&-
&-
&-\\
                &QQP&78.2
&84.0
&\bf 85.9
&-
&-
&-&&QQP&79.6
&85.9
&\bf 87.7
&-
&-
&-\\
                &MNLI&71.5
&78.3
&\bf 80.4 
&-
&-
&-&&MNLI&77.5
&83.7
&\bf 85.5
&-
&-
&-\\
                &RTE&62.5
&69.8
&\bf 71.9
&-
&-
&-&&RTE&70.3
&78.8
&\bf 80.7
&-
&-
&-\\
                &STS-B&79.0
&86.7
&\bf 88.5
&-
&-
&-&&STS-B&81.2
&87.9
&\bf 90.1
&-
&-
&-\\
                \midrule
                \multirow{5}{*}{\bf BERT-large}
                                &CoLA&53.4
&57.9
& 61.8
&62.0
&62.7
&\bf62.7&\multirow{5}{*}{\bf RoBERTa-large}
                                &CoLA&58.1
&59.7
& 62.0
&65.8
&66.9
&\bf67.8\\
                &QNLI&83.4
&84.7
& 88.5
&91.1
&91.1
&\bf91.1&&QNLI&85.2
&90.6
&93.8
&95.1
&95.1
&\bf95.1\\
                &QQP&80.6
&82.4
& 85.5
&86.9
&87.2
&\bf87.2&&QQP&82.4
&86.5
& 88.7
&90.3
&90.3
&\bf90.3\\
                &MNLI&73.7
&80.5
&82.3
&82.9
&82.9
&\bf83.6&&MNLI&79.6
&83.1
&85.4
&87.2
&89.5
&\bf89.5\\
                &RTE&65.8
&67.3
& 70.8
&73.1
&73.1
&\bf73.1&&RTE&72.1
&75.6
&80.5 
&83.0
&84.7
&\bf85.9\\
                &STS-B&82.3
&83.2
&86.4 
&89.5
&89.8
&\bf90.0&&STS-B&82.8
&84.3
& 88.4
&90.9
&91.9
&\bf91.9\\
                \bottomrule
            \end{tabular}}
        \end{threeparttable}
    \end{center}
    \label{tab:layer}
    % \vspace{-4mm}
\end{table*}

\textbf{The effect of each module. }
The results for the effect of each module are shown in Table~\ref{tab:com}. We first unfreeze any one module of the adapter tuning method (see row 2-5). The results represent that the bias vectors in the Hadamard adapter and the normalization module contribute more than the weight vectors (see row 2-5). We also unfreeze the normalization module (those right after intermediate outputs, short as normalization module) and attention-based normalization module (those right after self-attention outputs, short as attention-based normalization module), respectively. We find that the coarse-grained normalization modules (i.e. normalization module) are more necessary than the fine-grained normalization modules (i.e. attention-based normalization module) (see row 4-5).
Next, we unfreeze any two modules of the adapter tuning method (see row 6-10). The results represent that the most two effective modules, i.e., bias vectors in the Hadamard adapter and the normalization module, also achieve the best performance after being unfrozen simultaneously.
After that, we unfreeze three or four modules of the adapter tuning method (see row 11-13). Compared with the final three modules, we observe that when we add the attention-based normalization module, the performance decreases a little bit. It represents that some parameters are not valuable enough to improve PLMs' performance in the downstream tasks.

\textbf{The effect of the number of unfreezing layers. }
The results for the number of unfreezing layers of the Hadamard adapter on downstream tasks are shown in Table~\ref{tab:layer}. We select two tasks to visualize the trend as shown in Fig.~\ref{fig:layer}. We find that as we unfreeze more layers, the performance increase consistently. And they achieve satisfying performance when we unfreeze over a half of all layers (8 for models of base version and 16 for those of large version). The experiment inspires us that parameters in some layers of the Hadamard adapter are still redundant which can be removed to achieve more parameter-efficient with 0.022\% paramters, but this conjecture need to be validated with more datasets and PLMs.

\section{Exploratory analysis}
\begin{figure*}[!t]
  \centering
  \includegraphics[width=0.78\linewidth]{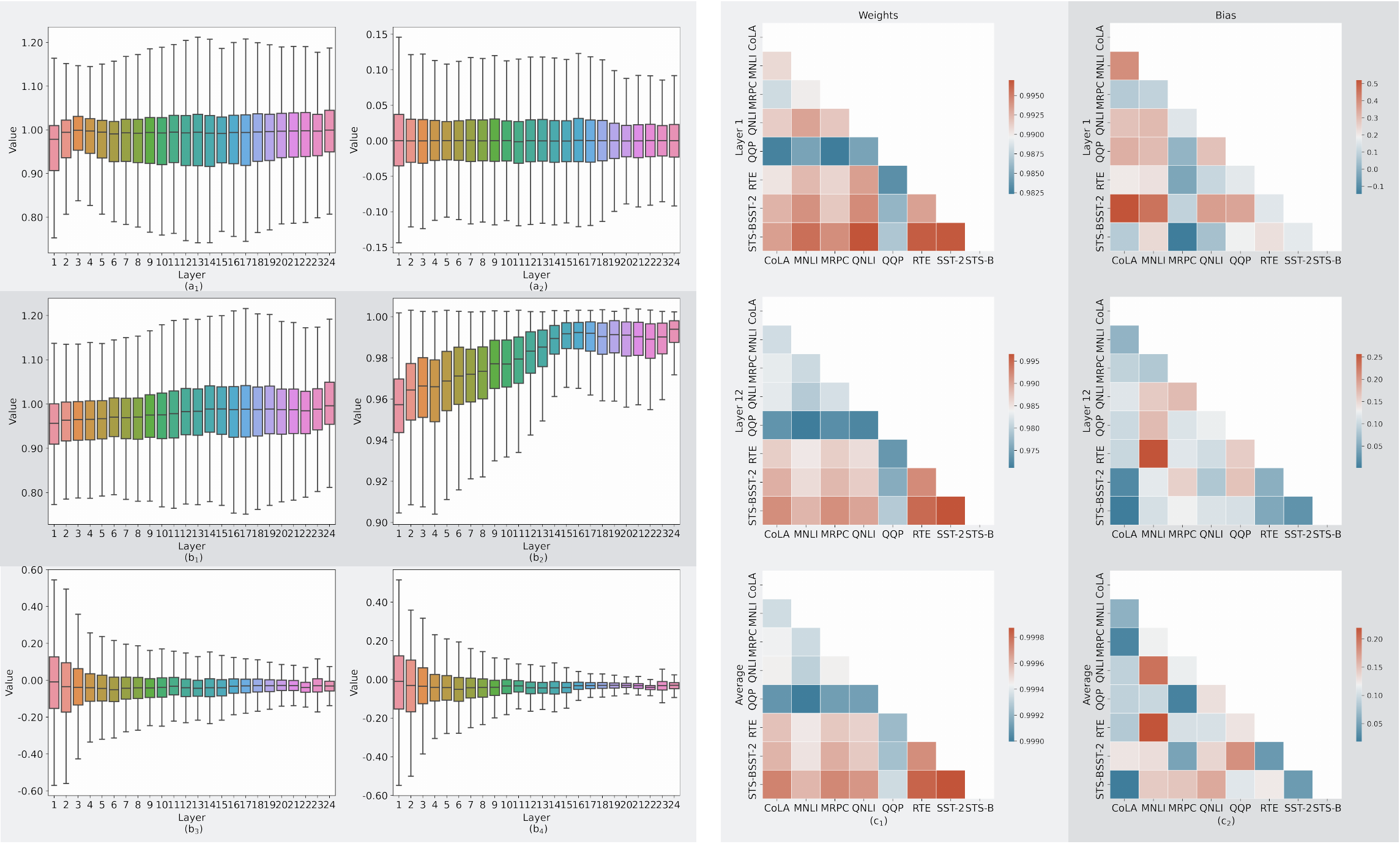}
  \caption{Each module in the Hadamard adapter based on each layer to answer three questions. Question one corresponds to (a$_1$) and (a$_2$), Question 1 corresponds to from (b$_1$) to (b$_4$), and Question 3 corresponds to (c$_1$) and (c$_2$)}
  \label{fig:pattern}
\end{figure*}

%pattern和task关联性，表征模型如何适应下游任务，可以作为描述下游任务特征的，发现不同下游任务pattern有共性。放在intro，有趣的发现
% To provide guidance for better applying Hadamard adapter to various downstream tasks, we conduct exploratory analysis to analyze the pattern of the learned adapters as shown in Fig~\ref{fig:pattern}.
% , and then analyze the patterns of the Hadamard adapter among different downstream tasks.
% In this section, we conduct exploratory analysis to analyze the pattern of the learned adapters as shown in Fig~\ref{fig:pattern}.
% For each downstream task, we could learn the Hadamard adapter, which represents the characteristics of the task. 
To get deeper understanding on the tuning results and provide guidance for better applying Hadamard adapter to various downstream tasks, 
we would like to visualize and analyze each module in the Hadamard adapter of each layer among different downstream tasks as shown in Fig~\ref{fig:pattern}. 

The overall analysis aims to answer the following three questions: 
i) For various downstream tasks, which layer of the adapter has the greater weight and bias variation? (Note that small variation presents the consistency among the learned adapters on different downstream tasks.)
ii) What is the difference between normalized module distribution under adapter tuning and full fine-tuning for various downstream tasks?
and iii) What are the commonalities of weight and bias in adapter between different downstream tasks?
Here we take RoBERTa-large model as an example to conduct the analysis, and the other PLMs are observed to have similar conclusions.

For the first question, we find that the weight and bias vectors of all datasets basically vary around 1.0 and 0.0 in each layer, respectively (see Fig~\ref{fig:pattern} (a$_1$)(a$_2$)). One box represents the distribution of the weight and bias vector values among all downstream tasks in the corresponding layer, respectively. in two sub-figures.
%
% The most volatile layers, that are layers with the largest variance, 
From the box plot, we find that the variance and extremum (maximum or minimum values) of weight and bias vectors in different layers are similar, respectively. 
%
% are the middle layers. While, the fewest ones are the front layers. Similarly, the layers with the most changes in maximum or minimum values is the middle layer, and the ones with the fewest changes are the front layers. 
%差异不大，
% It means the learned weight vectors of the adapters on different tasks are similar in front layers, while have big difference in middle layers. 
%
% For the bias vector, we find that all datasets basically vary around 0.0 in each layer (see Fig~\ref{fig:pattern} (a$_2$), one box represents the distribution of the bias vector values among all downstream tasks in the corresponding layer).
% %
% The most volatile layer is the first layer, and the smallest ones are the back layers. Similarly, the layers with the most changes in maximum or minimum values is the front layer, and the ones with the fewest changes are the back layers. 
%整体差异不大
The results represent that \emph{
the consistency degree of the learned Hadamard adapters among different tasks is similar in different layers.
% Hadamard adapter make great contributions in the front or middle layers. Those layers might be more necessary to be optimized to improve the performance of downstream tasks.
}
%不同任务之间一致性

For the second question, the weight vectors of the subsequent normalization module both vary around 1.0 after adapter tuning (see Fig~\ref{fig:pattern} (b$_1$), one box represents the distribution of the weight vector values among all downstream tasks in a layer after adapter tuning) and full fine-tuning (see Fig~\ref{fig:pattern} (b$_2$), one box represents the counterparts after full fine-tuning). 
In the adapter tuning, we also find that the variance and extremum of weight vectors in different layers are similar
%
% the most volatile layers, as well as the layers with the most changes in extremum are same as those of weight vectors in the Hadamard adapter (i.e. that is middle layers and front layers, respectively). 
%
Different from them, in the full fine-tuning, both the most volatile layers and the most changes in extremum are the front layers, and the corresponding fewest ones are both the back layers. 
Moreover, the bias vectors of the normalization module both vary around 0.0 after adapter tuning (see Fig~\ref{fig:pattern} (b$_3$), one box represents the distribution of the bias vector values among all downstream tasks in a layer after adapter tuning) and full fine-tuning (see Fig~\ref{fig:pattern} (b$_4$), one box represents the counterparts after full fine-tuning). 
We find that the trend of most volatile and extremum in adapter tuning and full fine-tuning are similar. The most volatile layers and the most changes in extremum are both the front layers, and the corresponding fewest ones are both the back layers. 
%
% They changes  same as those of weight vectors in full fine-tuning (i.e. that is front layers and back layers, respectively). 
%
The results represent that \emph{
% the normalization module plays an important role in the middle or front layers both in adapter tuning and full fine-tuning. Those layers might be more necessary to unify the distribution of self-attention outputs.
%不同任务的consistency
%后层consistency大于前层 
%weight consistency和bias consistency 不太一样
%ft参数
the trend of the consistency degree among different tasks along with layers in full fine-tuning and adapter tuning is general similar.
}

%举例子
For the third question, we calculate the cosine similarity of weight vectors (see Fig~\ref{fig:pattern} (c$_1$)) and bias vectors (see Fig~\ref{fig:pattern} (c$_2$)) in the Hadamard adapter, respectively, between each two downstream tasks. Red color represents more similar between weight or bias vectors of two tasks and blue color represents less similar. Due to limitation of space, we only display the heatmaps of the first, the middle layer and the average results. From the heatmaps, we find that the similarity of weight vectors in each layer for different tasks are almost same and consistent, and the values are closed to 1.0. Meanwhile, the similarity of bias vectors in each layer are obviously different, and achieve 0.3 at most.
%
% The results represent that \emph{bias contributes most in the Hadamard adapter, which inspires us that different tasks might share a common weight vectors during adapter tuning whose trainable parameters can be further reduced. But it needs more verification in the future study.}
%
The results indicate that bias contributes most in the Hadamard adapter and some adapter weights can be reused across different tasks, potentially leading to a more efficient and generalizable adapter tuning approach. The implications of this finding are significant, as it suggests that using a shared adapter approach could provide a more efficient and effective way to fine-tune pre-trained models for multiple tasks. By sharing weight vectors across tasks, the adapter network can be made smaller and less complex, reducing the risk of overfitting and improving the model's generalization performance.
% However, it is important to acknowledge that the applicability of this approach may depend on the specific tasks being considered and the degree of commonalities they share. 
Further research is required to explore the extent to which adapters can be shared across tasks.
% and to understand the potential benefits and limitations of such an approach.

\section{Related work}
% As the scale of Pretrained Language Models (PLMs) continues to increase, how to reduce the scale of parameters in fine-tuning PLMs for various downstream tasks becomes an increasingly important issue. 
Previous parameter-efficient fine-tuning of PLMs contains three main categories of methods, i.e., adapter tuning, prefix tuning, and prompt tuning.
%
%Parameter efficiency aims to use only a few parameters of plm to achieve full fine-tuning performance for downstream tasks. Reducing computational and storage cost of Pretrained Language Models (PLMs). 
%
%It aims at employ only a small amount of parameters of PLMs to achieve the performance of full fine-tuning on downstream tasks. The common parameter-efficient methods contain adapter tuning, prefix tuning, and prompt tuning, etc. 
%
Adapter tuning~\citep{Parameter-Efficient_Transfer_Learning} is to inject a small neural network module in each layer of PLMs, and only fine-tune parameters in this small neural network module. For instance, \citet{LORA} propose Low-Rank Adaptation (LoRA) which injects trainable rank decomposition matrices into each layer of the Transformer architecture. \citet{Parameter-efficient_Multi-task_Fine-tuning} learn adapter parameters for all layers and tasks by generating them using shared hypernetworks in a transformer model. \citet{liu2022few} introduce IA3, which incorporates three vectors (lk, lv, and lff) in each Transformer layer block, going beyond the simple addition of weight and bias vectors to the self-attention outputs. \citet{qi2022parameter} propose LN-tuning, which focuses on keeping only the gain term and bias term trainable in the LayerNorm module. 
Prefix tuning~\citep{Prefix-Tuning} and prompt tuning~\citep{The_Power_of} preset additional adjustable prefix tokens in the input layers or hidden layers, and only train these soft prompts during the fine-tuning on downstream tasks. For instance, \citet{GPT_Understands_Too} propose a prefix tuning method P-tuning which improve GPTs and BERTs performance in both few-shot and fully supervised settings.
\citet{Black-Box_Tuning_for} propose the
black-box tuning framework to optimize the continuous prompt prepended to the input text via derivative-free optimization.  
\citet{PPT} pre-train prompts by adding soft prompts into the pre-training stage to obtain a better initialization. 
\citet{he2021towards} propose a MAM Adapter which use prefix tuning with a small bottleneck dimension at the attention sub-layers.
% and allocates more parameter budgets to modify FFN representation using the scaled parallel adapter. 

%In addition to the above three parameter-efficient fine-tuning ways, the existing efforts also works on model compression~\citep{A_Survey_of}.

In addition to the above three parameter-efficient fine-tuning ways, there are also some other related research. For example, \citet{BitFit} propose a Bias-term
Fine-Tuning method which train only the bias-terms and the task-specific classification layer. \citet{Enabling_Lightweight_Fine-tuning} present a novel PLMs' compression approach based on the matrix product operator. 
\citet{Composable_Sparse_Fine-Tuning} propose Lottery Ticket Sparse Fine-Tuning conceived for pruning of large neural networks.
\citet{Y-Tuning} learns dense representations for labels Y when training PLMs and aligns them to fixed feature representation.
\citet{Raise_a_Child} only update a subset of parameters of PLMs via masking out the gradients of the non-child network during the backward process. 
~\citet{mao2021unipelt} introduce UNIPELT that learns to activate (upweight) the submodules that best suit the current task or specific data sample and deactivate (downweight) the rest. 
% Yang's work~\citep{yang2022parameter} operates on the raw hidden states, which is different from adapters and prompt tuning methods.

While the existing research significantly enhances the parameter efficiency when training PLMs, we hypothesize that there still exist superfluous parameters that could be eliminated. To address this, we introduce a highly efficient Hadamard adapter, which boasts the fewest parameters to date, yet delivers performance on par with full fine-tuning methods.

\section{Conclusions and future work}
%Pre-trained Language Models (PLMs) play important roles in Natural Language Processing (NLP). However, as we all know, most PLMs have billions of parameters, such as RoBERTa, T5, etc, which spend unaffordable computational resources and time to train. Therefore, constructing a parameter-efficient adapter is a valuable method to reduce parameters but do not damage the performance of PLMs on downstream tasks. 

%In this paper, we design a Hadamard adapter which only acts on self-attention outputs in PLMs. This novel adapter uses the element-wise linear transformation, which requires the fewest parameters compared to the previous parameter-efficient tuning methods.
%Besides, we also sum up some valuable patterns of the Hadamard adapter among downstream tasks, which is prepared for adopting shared adapters to further reduce parameters in the future study.
%
%The experiments conducted on the widely-used GLUE benchmark with several SOTA PLMs prove that the Hadamard adapter achieves competitive performance with only 0.033\% parameters compared with full fine-tuning, and it has the fewest parameters compared with other adapters. Moreover, we further find that there is also some redundant layers in the Hadamard adapter which can be removed to achieve more parameter efficiency with only 0.022\% parameters.

In this paper, we propose a Hadamard adapter based on a simple but effective element-wise linear transformation on the outputs of self-attention in PLMs.
We make comprehensive analysis for the feasibility of the Hadamard adapter, and summarize out some valuable patterns of it on downstream tasks. The experiments demonstrate that the proposed Hadamard adapter achieves competitive performance with 0.033\% parameters compared with full fine-tuning. Moreover, some layers in the Hadamard adapter are considered redundant to be removed for more parameter efficiency with 0.022\% parameters, which will be further investigated in the future study.

\section{Acknowledgement}
Some of computational resource are partially supported by 
Shanghai Municipal Science and Technology Major Project 

\noindent (No.2021SHZDZX0103), 
Science and Technology Commission of Shanghai Municipality Grant (No. 22511105902),
National Key Research and Development Project (No.2020AAA0109302), National Natural Science Foundation of China (No.62072323), Shanghai Science and Technology Innovation Action Plan (No. 22511104700, 22511105902), Shanghai Municipal Science and Technology Major Project (No.2021SHZDZX0103), and Science and Technology Commission of Shanghai Municipality Grant (No. 22511105902).

\clearpage

\bibliographystyle{ACM-Reference-Format}
\balance
\bibliography{main}

\end{document}